\documentclass[conference]{IEEEtran}
\IEEEoverridecommandlockouts

\usepackage{cite}
\usepackage{amsmath,amssymb,amsfonts}
\usepackage{algorithmic}
\usepackage{graphicx}
\usepackage{textcomp}
\usepackage{xcolor}
\usepackage[numbers,sort&compress]{natbib}
\def\BibTeX{{\rm B\kern-.05em{\sc i\kern-.025em b}\kern-.08em
    T\kern-.1667em\lower.7ex\hbox{E}\kern-.125emX}}
\begin{document}

\title{Opportunities and Challenges of Generative-AI in Finance 
\thanks{* These authors contributed equally.}
}

\makeatletter
\newcommand{\linebreakand}{%
  \end{@IEEEauthorhalign}
  \hfill\mbox{}\par
  \mbox{}\hfill\begin{@IEEEauthorhalign}
}
\makeatother

\author{\IEEEauthorblockN{Akshar Prabhu Desai \footnotemark{*}}
\IEEEauthorblockA{akshar.iitb@gmail.com}
\and
\IEEEauthorblockN{Tejasvi Ravi \footnotemark{*}}
\IEEEauthorblockA{ravitejasvi@gmail.com}
\and
\IEEEauthorblockN{Mohammad Luqman}
\IEEEauthorblockA{luqmankgp@gmail.com}
\linebreakand
\IEEEauthorblockN{Ganesh Mallya}
\IEEEauthorblockA{mallya.sg@gmail.com}
\and
\IEEEauthorblockN{Nithya Kota}
\IEEEauthorblockA{nithyakota@google.com}
\and
\IEEEauthorblockN{Pranjul Yadav }
\IEEEauthorblockA{ipranjul@gmail.com}
\linebreakand
\IEEEauthorblockA{Google}
}
\maketitle

\begin{abstract}

Gen-AI techniques are able to improve understanding of context and nuances in language modeling, translation between languages, handle large volumes of data, provide fast, low-latency responses and can be fine-tuned for various tasks and domains. 

In this manuscript, we present a comprehensive overview of the applications of Gen-AI techniques in the finance domain. In particular, we present the opportunities and challenges associated with the usage of Gen-AI techniques. We also illustrate the various methodologies which can be used to train Gen-AI techniques and present the various application areas of Gen-AI technologies in the finance ecosystem. 

To the best of our knowledge, this work represents the most comprehensive summarization of Gen-AI techniques within the financial domain. The analysis is designed for a deep overview of areas marked for substantial advancement while simultaneously pin-point those warranting future prioritization. We also hope that this work would serve as a conduit between finance and other domains, thus fostering the cross-pollination of innovative concepts and practices.

\end{abstract}


\section{Introduction}
Financial sector has historically applied traditional machine learning technologies to address domain specific challenges. For example, clustering \cite{cluster}, predictive analysis \cite{prediction}, anomaly detection \cite{anomaly}, time-series monitoring \cite{timeseries}, and graph-based learning have found extensive use for finance-specific problems such as detecting trends, prediction, risk modeling, and better data representation. However, Generative AI (Gen-AI) technologies, such as large language models (LLMs), have a superior ability to understand tasks that involve language compared to any traditional techniques. Kashyap et al. \cite{kashyap} discuss that LLMs can effectively perform traditional machine learning tasks and, when used correctly, can even outperform traditional methods such as classification.

Gen-AI techniques are able to improve understanding of context and nuances in language modeling \cite{10.1145/3649506}, translation between languages, handle large volumes of data, provide fast and low-latency responses and can be fine-tuned \cite{xia2024understanding} for various tasks and domains. The advancement was sparked by the development of transformer architecture, which is based solely on attention mechanisms, dispensing with recurrence and convolutions \cite{attention}. Experiments have been performed to demonstrate how Gen-AI techniques have superior quality while requiring significantly less time to train \cite{tan2021efficientnetv2, sun2024cebench}. 

There are several opportunities associated with the usage of Gen-AI technologies in finance \cite{aifintech}.  Broadly speaking, they can be used across multiple avenues such as interactive (e.g. chatbots), assistive (e.g. easing payment related activities, right card to use), educative (e.g. enabling users to understand finance concepts) and advisory (e.g. trading assistant).

Along with the opportunities, there lies several challenges associated with the widespread adoption. Some of these challenges originate from the scarcity of private and sensitive quality data available for training, issues stemming from pre-training and fine-tuning, inference latency, computational costs associated with deploying these models in production, exorbitant cost associated with the pricing of APIs available for commercial usage and the inherent biases embedded within these models \cite{shabsigh2023generative, singh2024generative}. 

There are several avenues via which Gen-AI technologies can be trained and developed for financial use-case. Users can leverage either open source models or external service providers for their tasks \cite{zhao2024revolutionizingfinancellmsoverview}. Alternatively, one can explore the usage via prompt engineering i.e. zero-shot and few-shot learning. In scenarios, when utilizing Gen-AI techniques out of the box is not working for the task at hand, then one could leverage fine-tuning (e.g. instruction specific, task-specific or parameter efficient) to train in order to perform better at the specific tasks the user has in mind. 

Gen-AI techniques also been widely applied in the finance industry with its application to customer service and support (e.g. sentiment analysis, chat-bots), text summarization and assistance (e.g. summaries from large text, recommending knowledge from large corpora), auto-filling forms, risk management (e.g. market risk analysis, credit risk monitoring, anomaly detection), investment trading (e.g. quantitative analysis) and document processing (e.g.  regulatory compliance) \cite{cao2024empowering} \cite{goldberg2024ai}.

The organization of this paper is as follows. In Section \ref{sec:opportunity}, we discuss the various opportunities associated with the usage of Gen-AI in the finance ecosystem. In Section \ref{sec:challenge}, we present the various challenges associated with the usage of these techniques. Further, in Section \ref{sec:methodologies}, we illustrate the various methodologies which can be used to train  Gen-AI for widespread adoption. Furthermore, in Section \ref{sec:applications}, we present the various application areas of Gen-AI in finance. Lastly, in Section \ref{sec:concluion}, we conclude this manuscript.

\section{Opportunities of Gen-AI in finance} \label{sec:opportunity}
In this section, we outline the opportunities associated with the usage of Gen-AI techniques in the finance domain. Broadly, we classify the opportunities into four major categories i.e. interactive, assistive, educative and advisory.

\subsection{Interactive}
Gen-AI technology has the potential to become a state-of-the-art technology for both Task-Oriented Dialog  \cite{survey-of-tod} and Open-Domain Dialog, thereby making it the go-to technology for building conversational and interactive applications. The associated inherent ability with these techniques to ask follow up questions, track references to previous entities, and user preferences makes them more coherent and consistent. They are a vast improvement over rule-based chat-bots which could respond to a fixed set of queries with a pre-configured set of responses. Fine-tuning and Retrieval Augmented Generation (RAG) \cite{rag} can further help Gen-AI techniques, become more domain specific and/or access more up to date relevant information to support customers ranging from routine to complex inquiries while creating more personalized and engaging experiences. McKinsey \cite{gaiPotential} estimates that Gen-AI could further reduce the volume of human-serviced contacts by up to 50 percent, depending on a company’s existing level of automation. Further, Klarna, a Buy Now Pay Later (BNPL) company has observed that using Gen-AI techniques within its customer service tool can lead to a \$40M impact \cite{KlarnaInternationalPress}.

Gen-AI techniques could also be used to create appealing backgrounds for debit, credit cards, coupons and gift cards. Many models including Parti \cite{Parti} can generate high-fidelity photo-realistic images and supports content-rich synthesis involving complex compositions and world knowledge. Further, Gen-AI techniques can be used to create dynamic visually pleasing themes for apps that are more engaging and context specific.



\subsection{Assistive}
Gen-AI techniques have the ability to assist through digital automation, co-piloting, auto-filling and  shopping.

In particular, agents using Gen-AI techniques, can be trained to execute APIs \cite{schick2023toolformerlanguagemodelsteach} thereby enabling them to execute tasks, automate routines thereby substantially improving the quality of life with humans having mobility and dexterity challenges. Further, a payments co-pilot can help identify opportunities for better payment options among cards, instrument modes (e.g. cards, bank accounts).  

Gen-AI techniques are able to parse and understand page content and hence can help towards auto-filling, which is a complex problem due to the dynamic and changing nature of forms across the internet. Further, agents using Gen-AI technology can be used as digital assistants. LLaSA \cite{zhang2024llasa}, a digital assistant is able to parse the product inventory and provide recommendations that closely match the user’s needs. 

\subsection{Educative}
Gen-AI techniques have the potential to create applications which can guide, educate and empower users w.r.t. financial literacy.

Gen-AI technology can be used to perform deep dives into existing financial data about a customer or company, thereby uncovering financial trends that could lead to cost efficiency measures, or investing opportunities. In particular, Gen-AI techniques have the ability to ingest vast amounts of financial data to analyze market sentiment, news and other data sources to help inform/create better trading strategies \cite{aimlfin}. 

Further, chatbots \cite{chatbot} built using Gen-AI techniques can understand complex topics like mortgage planning and investments strategies (e.g. stock options, mutual funds).

\subsection{Advisory}
Financial institutions can leverage the capabilities of Gen-AI techniques to build powerful tools for financial advisory applications (e.g. risk assessment, pro-active detection, summarization and recommendation). 

Gen-AI techniques can be used to build sophisticated machine learning models by processing vast amounts of financial data to identify anomalous patterns of fraud. Such models can be used for better credit scoring,  risk assessment for loans, to identify fraudulent activities in payment systems and to forecast financial risks using chain of thought prompting \cite{chainofthought} to elicit better reasoning and accuracy. Further, Gen-AI techniques have demonstrated a remarkable ability towards document summarization (i.e. concise yet informative) and identifying key themes.


\section{Challenges of using Gen-AI in finance} \label{sec:challenge}

Strict regulations and constraints within the financial domain hinders the development of Gen-AI applications. Further, Gen-AI's existing limitations present challenges for effective use in financial applications. In this section, we explore these challenges in more detail.

\subsection{Data}
\subsubsection{Data Constraints}
There is a dearth of publicly available financial datasets owing to the sensitive nature of financial data. This limits the availability of open financial Gen-AI models \cite{maple2023ai}. 
BloombergGPT, a proprietary Gen-AI model, is the only major model trained for financial tasks from ground up. BloombergGPT is trained on 363 billion tokens of Bloomberg's own proprietary data and 345 billion tokens of publicly available data \cite{BloombergGPT}.

While many financial institutions have access to proprietary data, they still face data availability challenges due to the institutional inertia and privacy related concerns. Kruse et al. \cite{kruse2019artificial} conducted a survey of financial service experts to identify key challenges of using Gen-AI and two thirds of the experts indicated the lack of quality training data as the primary obstacle. Further, the same study \cite{kruse2019artificial} identified financial institutions' unwillingness to move the data to cloud as one of the major limiting factors in using Gen-AI in finance. Financial institutions see their data as their core-business and have concerns in moving this data to distributed data centers (or cloud) which is necessary for training. 

The lack of quality training data and privacy concerns can be addressed using synthetic data \cite{assefa2020generating}. Synthetic data is artificially generated data that mimics the statistical properties of the real world data. Since the data is artificially generated, it also helps alleviate the concerns around data privacy. Samuel et al. \cite{assefa2020generating} provides a detailed overview of the current state of art of synthetic data generation in finance. However, they also highlight the challenges with synthetic data such as over-fitting and lack of open benchmarks. 

\subsubsection{Nature of Financial Data}
Ljung et al. \cite{ljung2021synthetic} showed that financial data is different from data in other domains because it does not follow Gaussian distribution and hence can not be normalized using state-of-the-art approaches for downstream modeling tasks. This requires the creation of new pre-processing methods for machine learning use-cases.
Further, Samuel et al. \cite{assefa2020generating} found that multi-modality and heterogeneity of financial data along with its intricate inter-dependencies are harder to be captured and modeled by Gen-AI technologies. 

\subsection{Challenges in fine tuning} \label{challengefinetune}
Gen-AI models require substantial fine tuning efforts when they are applied to specific domains such as finance \cite{devlin2019bertpretrainingdeepbidirectional}. Fine-tuning is a technique to adapt a pre-trained model for a more specific purpose (e.g. financial applications). Li et al. \cite{li2023large} provided a comprehensive comparison of large language models (e.g. LLaMa) for financial applications and observed that fine tuned models can outperform generic models w.r.t financial tasks.

Further, FinBERT \cite{liu2021finbert}, which uses LoRA (Low Rank Adaptation) technique \cite{lora} to fine-tune open source LLMs, highlights the importance of pre-training along with fine tuning. In particular, they emphasized that financial vocabulary needs to be carefully embedded into the pre-training stages along with fine-tuning (e.g., associating “bank” with the word “lending” as compared to the word “river”).

Li et al. \cite{li2023large} also highlights the critical nature of high quality human labeled data for fine tuning. They mentioned that, in the finance domain, cost of producing human labeled data is harder and expensive due to challenges around regulations, privacy and availability of humans with specialized knowledge.  

\subsection{Computation costs}

The two major contributors to computational costs are training cost and inference cost. 
Computational costs associated with training Gen-AI models from the ground up can easily run into billions of dollars \cite{xia2024understandingperformanceestimatingcost}. However, Xia et al. \cite{xia2024understandingperformanceestimatingcost} noted that fine tuning \ref{challengefinetune} can help reduce this cost for certain applications. 

Inference cost associated with the integration of Gen-AI workflows will be high considering the billions of stock market trades, credit card payment and banking transactions within a day \cite{HANCOCK19971573}. Further, pricing models of Gen-AI APIs available for commercial use vary from \$5 to \$15 per million input tokens. For example, if we assume that each credit card transaction  involves 500 tokens, this translates to a cost of \$0.0025 per transaction or \$5 million for 2 billion transactions per day in additional expenses. Bryce et al. \cite{brycetrends} provided an overview about how these pricing costs would impact Gen-AI deployment in finance.  


\subsection{Secondary Considerations}
Additional considerations of using Gen-AI techniques in finance domain can arise from domain independent issues, embedded domain bias, privacy and regulatory concerns.

Domain independent issues of Gen-AI techniques such as hallucination and inconsistent reasoning \cite{srivastava2024bai} using incorrect information might lead to significant financial loss \cite{10.1145/3604237.3626867}.

Embedded domain bias stems from the fact that different regions have different laws governing the kind of information, which can be used for making financial decisions (e.g. credit worthiness) \cite{glavina2024ai}. As a result, Gen-AI techniques trained on such data might further perpetuate biases without being fully aware of their existence.

Data leaks due to sophisticated prompt engineering techniques \cite{huang2024genai} raises substantial privacy and regulatory concerns \cite{shabsigh2023generative, zhang2023right} surrounding users private data. This complexity further increases due to new privacy regulations (e.g. right to be forgotten) adopted in many regions of Europe \cite{rosen2011right}. This increase in complexity can be attributed to the fact that a model might still preserve some aspects of the data as part of its learning, even though the underlying data is requested to be deleted.

\section{Gen-AI Methodologies} \label{sec:methodologies}

This section presents various ways to train or fine-tune Gen-AI techniques for potential use-cases in finance.

\subsection{Out-of-Box}

In this technique, the users leverage either open source models or Gen-AI service providers like OpenAI, Gemini, Microsoft and Perplexity for their tasks. This method requires no training data, and either minimal to no compute resources. 

Further, users can make use of prompt engineering \cite{prompt} techniques to complete their task. Broadly, there are two main varieties of prompt engineering techniques i.e. zero-shot \cite{zeroshot} and few-shot \cite{brown2020language}. In the case of zero-shot learning, Gen-AI techniques are able to perform the task at hand without seeing any prior examples, but just by leveraging the knowledge present in them. On the other hand in case of few-shot learning, users provide a few examples as part of the prompt for the Gen-AI techniques to learn and perform the task at the end of the prompt. In particular, in this technique, the Gen-AI techniques can leverage their induction heads \cite{inductionheads} to perform in-context learning \cite{incontextlearning} and arrive at the result.

\subsection{Fine-tuning}
In scenarios, where utilizing LLMs out of the box is not working for the task at hand, then users could leverage fine-tuning to train the LLMs to perform better at the specific tasks the user has in mind. At a high level, fine-tuning techniques can be divided into the following categories.

\subsubsection{Instruction Fine-tuning} \label{instructionfinetuning}
In instruction fine-tuning \cite{finetuning}, the pre-trained LLMs are further trained on a labeled set of prompts and answer pairs. This newly trained LLM is tuned to answer specifically to specific kinds of instruction/prompt. As the name suggests, the training data need to be in the form of instructions. So users need to either collect the training data in the form of instructions or could leverage prompt template libraries like prompt-engine-py or dynamic prompts to take normal datasets and convert them into instruction datasets for fine-tuning.

\subsubsection{Task specific Fine-tuning} \label{taskspecificfinetuning}
This technique \cite{tasktuning} involves the users fine-tuning a pre-trained Gen-AI model to perform a specific kind of task in mind. For example, the users might fine-tune the pre-trained LLM to perform sentiment detection given an input prompt. It involves very few examples for the LLM to train on, but still requires a decent amount of compute as the entire model needs to be loaded into memory for the training part.

Task specific fine-tuning \cite{tasktuning} is prone to exhibit a phenomenon called catastrophic forgetting. In catastrophic forgetting, the underlying LLM has forgotten the knowledge of the world it had obtained as part of its pre-training and its performance on the other tasks after task specific fine-tuning is much worse than its performance on the same tasks before fine-tuning. In order to circumvent the issue of catastrophic forgetting, users can employ multi-task instruction fine-tuning or employ parameter efficient fine-tuning described in section \ref{peft}.
\subsubsection{Parameter Efficient Fine-tuning} \label{peft}
Instruction Fine-Tuning \ref{instructionfinetuning} or Task specific fine-tuning \ref{taskspecificfinetuning} are resource intensive and are plagued with catastrophic forgetting. In order to circumvent both these issues users can leverage Parameter Efficient Fine-tuning (PEFT) \cite{paramtuning}. Two most commonly used techniques in PEFT are LoRA \cite{lora} which rely on re-parametrization technique and soft prompts which add additional trainable layers to the LLM.

\begin{enumerate}
    \item LoRA introduces a new way to train the LLMs by retaining their pre-training knowledge. The weights of the LLM are frozen and new low rank decomposition matrices are added to every layer in the transformer. 
    \item Soft Prompts \cite{DBLP:journals/corr/abs-2104-08691} fall under the additive method paradigm where no weights of the model are changed. Instead of modifying the weights of the pre-trained LLMs, soft prompts rely on prepending trainable tokens or soft prompts to the input tokens during training for a given task. The loss is propagated all the way to these trainable tokens and an efficient representation is learnt for the task at hand. The idea of these soft tokens is to choose the right vectors for a given space on the N-dimensional hypersphere of the embedding vector space. Since these soft tokens are very light-weight, multiple such soft tokens can be learnt similar to LoRA.
\end{enumerate}

\subsection{Agentic Systems}
Even though LLMs have made great strides in their ability to understand natural languages (via zero-shot or few shot examples), LLMs still struggle in providing accurate up to date information, hallucinate answers \cite{xu2024hallucination, llmfactuality} or are unable to perform precise mathematical calculations \cite{arellmgoodatmath}. A very simple solution is to provide the ability for LLMs to utilize external tools such as search engine, calculators etc which help the LLMs in overcoming their major drawbacks.

\subsection{Quantization}
Traditionally models were trained and deployed using 32-bit floating point numbers to represent the parameters of the model. With the exponential increase in the number of parameters in LLMs, one major concern that arises is the amount of time taken for inference. In order to speed up the inference time researchers leverage quantization to represent the parameters of the model using less bits (float16, bfloat16 or int8) without much loss in accuracy. Users utilize one of the two quantization schemes
\begin{enumerate}
    \item Post Training Quantization: In this method a pre-trained 32-bit floating point number based model is converted into low bit numbers. The quantization maybe data free or a very small amount of data called calibration data can be used. A crucial step in this quantization scheme is to find a good quantization ranges for the quantizer as noted in \cite{DBLP:journals/corr/abs-2106-08295}. In this quantization scheme no retraining of the LLM is involved. This method of quantization (specifically the case with data free quantization) is particularly helpful when security or data privacy may limit data access \cite{DBLP:journals/corr/abs-2103-13630}.
    \item Quantization aware training (QAT) involves in retraining the pre-trained LLM using training data. In this quantization scheme since the quantization operation is non-differentiable, during back-propagation the gradient is approximated using an identity function also known as Straight Through Estimator \cite{DBLP:journals/corr/abs-1305-2982}. This method is particularly useful if the quantized model will be in use for a quite some time. Since this method involves retraining of the LLM on lower precision, the training must be performed for an extended period so that quantized LLM converges to better loss in the loss manifold \cite{DBLP:journals/corr/abs-2103-13630}.
\end{enumerate}
\section{Gen-AI Applications in finance ecosystem} \label{sec:applications}
Gen-AI techniques are still in its nascent stages and its implementation within finance sector comes with its own set of risks and challenges. However, the transformative nature of this technology is driving many applications in this domain. In this section, we will highlight such applications both in the industry and the research community.

\subsection{Numerical Reasoning}
Several datasets consisting of complex numerical reasoning tasks have been proposed to evaluate the efficacy of Gen-AI techniques. For example, FinQA \cite{finqa}, is a dataset with  Question-Answer pairs over Financial reports written by financial experts. FinQA, consist of questions such as “Considering the weighted average fair value of options, what was the change of shares vested from 2005 to 2006?”. Similarly, ConvFinQA \cite{convfinqa}, an extension of FinQA, is a multi-turn conversational question-answering dataset over financial reports, consisting of 3,892 conversations with 14,115 questions. GPT-4 achieved an accuracy of 78\% on FinQA and 76\% on ConvFinQA dataset which is much higher than the average human \cite{gpt-on-fin-datasets}. This ability has enabled real world used cases like Fintool \cite{fintool} - an AI equity research tool that is engineered to discover financial insights about companies.

Credit Karma \cite{graciano2024credit}, is making use of the Intuit Assist to launch Gen-AI experiences which include asking questions about a user's personal finance, understanding spend and personal financial roadmap.


\subsection {Trading}
Gen-AI techniques have also seen applications in trading yielding impressive results in research settings. In particular, Wu et al. (2023) introduced BloombergGPT \cite{BloombergGPT}, a 50-billion parameter Gen-AI technique, designed specifically for the financial domain. They demonstrated the effectiveness of their proposed technique for sentiment analysis, financial question answering, while maintaining proficiency in general language tasks.

Mai (2024) proposed StockGPT \cite{stockgpt}, which uses token sequences to learn predictive patterns via the attention mechanism there by demonstrating substantial impact. Further, Lopez et al. \cite{Lopez_Lira_2023} showcased that incorporating advanced Gen-AI techniques into the investment decision-making process can yield accurate predictions and enhance the performance of quantitative trading strategies. Furthermore, Fatourosa et al. \cite{can-llm-beat-wallstreet} proposed that a portfolio rebalanced monthly using the buy/sell signals generated from Gen-AI technique can outperform a passive index by 10-30\%. 

Additionally, Lezhi et al. \cite{li2023multimodal} showcased how multimodal Gen-AI techniques could be used for fundamental investment research. Through fine-tuning methods applied to a base model (Llama2), they developed an AI (Artificial Intelligence) agent that can assist investors in tasks such as understanding market conditions, generating investment ideas, and formatting results with stock recommendations.

\subsection{Summarization and Assistive Applications}

Gen-AI techniques have also been applied in text summarization and assistive applications. In particular,  Xiao et al. \cite{summarization-is-dead} observed a clear preference among human evaluators for LLM-generated summaries over human-written summaries. Gen-AI techniques like SaulLM-7B \cite{colombo2024saullm7b} based on mistral 7B architecture, are designed for legal text generation and comprehension.

Further, JP Morgan uses a Gen-AI toolkit, designed to serve as a ‘research analyst’, aiding in various tasks that enhance productivity and decision-making within the firm \cite{jpmorganResearch}. Furthermore, companies like Kudos and Max-Rewards understand credit card offerings and combine them with user's spends to suggest the best credit cards for maximising rewards. Morgan Stanley's COIN (Contract Intelligence) uses AI to review legal documents and extract relevant information. 

Leading CRM companies such has Salesforce \cite{salesforceEinsteinAI}, Hubspot \cite{hubspotChatbotBuilder}, Zendesk \cite{zendeskChatbotBuilder} offer AI based chatbots for better customer experience. Further, Intuit has integrated, a Gen-AI powered financial assistant called intuit assist into TurboTax \cite{turbotax}, that uses a combination of the company’s prior tax preparation experience, data and documents from the filer, and current tax code for tax filing. 

\subsection{Risk Monitoring}
Mastercard \cite{brown2024decision} discussed how Gen-AI techniques could help financial institutions improve their fraud detection rates by 20\%, on average. Further, Cao et al. \cite{risklabs} explored a new framework that leverages Gen-AI to analyze and predict financial risks. It uniquely combines different types of financial data, including textual and vocal information from Earnings Conference Calls (ECCs), market-related time series data, and contextual news data and showcases the critical role of LLMs in financial risk assessment and opens new avenues for their application in this field.

\section{Conclusion} \label{sec:concluion}
 
Compared to the traditional machine learning and data mining approaches, Gen-AI techniques are able to improve understanding of context and nuances in language modeling, translation between languages, handle large volumes of data, provide fast and low-latency responses and can be fine-tuned for various tasks and domains. In this manuscript we have provided a brief overview of the application of Gen-AI techniques within the finance ecosystem.

Further, we discussed the various opportunities and challenges associated with the widespread adoption of Gen-AI techniques in the finance domain. We also discuss techniques which can be used to develop Gen-AI models by overcoming challenges stemming from critical issues such as hallucination, data-quality and computational costs. Lastly, we provide application areas in Finance where these Gen-AI models are or can be used.  

To the best of our knowledge, this is the first work which comprehensively summarizes the usage of Gen-AI techniques in the finance domain. This summarization would not only provide a clear overview of the areas where substantial work has been completed but also pinpoint areas where future efforts can be prioritized, potentially accelerating the development and adoption of Gen-AI techniques in finance. Moreover, this analysis could serve as a valuable bridge between finance and other domains, facilitating the cross-pollination of innovative ideas and best practices, ultimately benefiting a wider range of industries and applications. 

\bibliography{paper.bib}

\begin{thebibliography}{77}
\providecommand{\natexlab}[1]{#1}
\providecommand{\url}[1]{\texttt{#1}}
\expandafter\ifx\csname urlstyle\endcsname\relax
  \providecommand{\doi}[1]{doi: #1}\else
  \providecommand{\doi}{doi: \begingroup \urlstyle{rm}\Url}\fi

\bibitem[Rokach and Maimon(2005)]{cluster}
Lior Rokach and Oded Maimon.
\newblock Clustering methods.
\newblock In \emph{Data mining and knowledge discovery handbook}, pages 321--352. 2005.

\bibitem[Aitchison and Dunsmore(1975)]{prediction}
John Aitchison and Ian~Robert Dunsmore.
\newblock \emph{Statistical prediction analysis}.
\newblock 1975.

\bibitem[Chandola et~al.(2009)Chandola, Banerjee, and Kumar]{anomaly}
Varun Chandola, Arindam Banerjee, and Vipin Kumar.
\newblock Anomaly detection: A survey.
\newblock \emph{ACM computing surveys (CSUR)}, 41\penalty0 (3):\penalty0 1--58, 2009.

\bibitem[Hamilton(2020)]{timeseries}
James~D. Hamilton.
\newblock \emph{Time series analysis}.
\newblock Princeton university press, 2020.

\bibitem[Kashyap and Sinha(2024)]{kashyap}
Y.~Kashyap and A.~Sinha.
\newblock Llm is all you need: How do llms perform on prediction and classification using historical data.
\newblock \emph{International Journal For Multidisciplinary Research}, 6\penalty0 (3), Jul 2024.

\bibitem[Yang(2024)]{10.1145/3649506}
J.~et~al. Yang.
\newblock Harnessing the power of {LLMs} in practice: A survey on chatgpt and beyond.
\newblock \emph{{ACM} Trans. Knowl. Discov. Data}, 18\penalty0 (6):\penalty0 1--32, Jul 2024.
\newblock \doi{10.1145/3649506}.

\bibitem[Xia(2024{\natexlab{a}})]{xia2024understanding}
Y.~et~al. Xia.
\newblock Understanding the performance and estimating the cost of llm fine-tuning.
\newblock \emph{arXiv:2408.04693}, 2024{\natexlab{a}}.

\bibitem[Vaswani(2017)]{attention}
Ashish et~al. Vaswani.
\newblock Attention is all you need.
\newblock \emph{arXiv preprint arXiv:1706.03762}, 2017.

\bibitem[Tan and Le(2021)]{tan2021efficientnetv2}
M.~Tan and Q.~Le.
\newblock {Efficientnetv2}: Smaller models and faster training.
\newblock In \emph{Proc. Int. Conf. Mach. Learn.}, pages 10096--10106, 2021.

\bibitem[Sun(2024)]{sun2024cebench}
W.~et~al. Sun.
\newblock {CEBench}: A benchmarking toolkit for the cost-effectiveness of llm pipelines.
\newblock \emph{arXiv:2407.12797}, 2024.

\bibitem[Milana and Ashta(2021)]{aifintech}
C.~Milana and A.~Ashta.
\newblock Artificial intelligence techniques in finance and financial markets: A survey of the literature.
\newblock \emph{Strategic Change}, 30\penalty0 (3):\penalty0 189--209, 2021.

\bibitem[Shabsigh and Boukherouaa(2023)]{shabsigh2023generative}
G.~Shabsigh and E.~Boukherouaa.
\newblock \emph{Generative Artificial Intelligence in Finance: Risk Considerations}.
\newblock International Monetary Fund, 2023.

\bibitem[Singh(2024)]{singh2024generative}
B.~Singh.
\newblock Generative artificial intelligence: Prospects for banking industry.
\newblock \emph{International Journal of Research in Engineering, Science and Management}, 7\penalty0 (3):\penalty0 83--86, 2024.

\bibitem[Zhao(2024)]{zhao2024revolutionizingfinancellmsoverview}
H.~et~al. Zhao.
\newblock Revolutionizing finance with llms: An overview of applications and insights.
\newblock \emph{arXiv:2401.11641 [cs.CL]}, 2024.

\bibitem[Cao(2024)]{cao2024empowering}
X.~et~al. Cao.
\newblock Empowering financial futures: Large language models in the modern financial landscape.
\newblock \emph{EAI Endorsed Transactions on AI and Robotics}, 3, 2024.

\bibitem[Goldberg(2024)]{goldberg2024ai}
A.~Goldberg.
\newblock Ai in finance: Leveraging large language models for enhanced decision-making and risk management.
\newblock \emph{Social Science Journal for Advanced Research}, 4\penalty0 (4):\penalty0 33--40, 2024.

\bibitem[Qin(2023)]{survey-of-tod}
Libo et~al. Qin.
\newblock End-to-end task-oriented dialogue: A survey of tasks, methods, and future directions.
\newblock \emph{arXiv preprint arXiv:2311.09008}, 2023.

\bibitem[Lewis et~al.(2020)Lewis, Perez, Piktus, Petroni, Karpukhin, Goyal, K{\"u}ttler, Lewis, Yih, Rockt{\"a}schel, et~al.]{rag}
Patrick Lewis, Ethan Perez, Aleksandra Piktus, Fabio Petroni, Vladimir Karpukhin, Naman Goyal, Heinrich K{\"u}ttler, Mike Lewis, Wen-tau Yih, Tim Rockt{\"a}schel, et~al.
\newblock Retrieval-augmented generation for knowledge-intensive nlp tasks.
\newblock \emph{Advances in Neural Information Processing Systems}, 33:\penalty0 9459--9474, 2020.

\bibitem[gai()]{gaiPotential}
The economic potential of generative ai: The next productivity frontier.

\bibitem[Kla()]{KlarnaInternationalPress}
Klarna ai assistant handles two-thirds of customer service chats in its first month.
\newblock Accessed: 2024-10-27.

\bibitem[Yu et~al.(2022)Yu, Xu, Koh, Luong, Baid, Wang, Vasudevan, Ku, Yang, Ayan, et~al.]{Parti}
Jiahui Yu, Yuanzhong Xu, Jing~Yu Koh, Thang Luong, Gunjan Baid, Zirui Wang, Vijay Vasudevan, Alexander Ku, Yinfei Yang, Burcu~Karagol Ayan, et~al.
\newblock Scaling autoregressive models for content-rich text-to-image generation.
\newblock \emph{arXiv preprint arXiv:2206.10789}, 2\penalty0 (3):\penalty0 5, 2022.

\bibitem[Schick et~al.(2023)Schick, Dwivedi-Yu, Dess`ı, Raileanu, Lomeli, Zettlemoyer, Cancedda, and Scialom]{schick2023toolformerlanguagemodelsteach}
T.~Schick, J.~Dwivedi-Yu, R.~Dess`ı, R.~Raileanu, M.~Lomeli, L.~Zettlemoyer, N.~Cancedda, and T.~Scialom.
\newblock Toolformer: Language models can teach themselves to use tools.
\newblock 2023.

\bibitem[Zhang(2024)]{zhang2024llasa}
S.~et~al. Zhang.
\newblock Llasa: Large language and e-commerce shopping assistant.
\newblock \emph{arXiv:2408.02006}, 2024.

\bibitem[Ahmed et~al.(2022)Ahmed, Alshater, El~Ammari, and Hammami]{aimlfin}
Shamima Ahmed, Muneer~M. Alshater, Anis El~Ammari, and Helmi Hammami.
\newblock Artificial intelligence and machine learning in finance: A bibliometric review.
\newblock \emph{Research in International Business and Finance}, 61:\penalty0 101646, 2022.
\newblock ISSN 0275-5319.
\newblock \doi{https://doi.org/10.1016/j.ribaf.2022.101646}.

\bibitem[Kim et~al.(2023)Kim, Chua, Rickard, and Lorenzo]{chatbot}
Jin~K Kim, Michael Chua, Mandy Rickard, and Armando Lorenzo.
\newblock Chatgpt and large language model (llm) chatbots: The current state of acceptability and a proposal for guidelines on utilization in academic medicine.
\newblock \emph{Journal of Pediatric Urology}, 19\penalty0 (5):\penalty0 598--604, 2023.

\bibitem[Wei et~al.(2022)Wei, Wang, Schuurmans, Bosma, Ichter, Xia, Chi, Le, and Zhou]{chainofthought}
Jason Wei, Xuezhi Wang, Dale Schuurmans, Maarten Bosma, Brian Ichter, Fei Xia, Ed~Chi, Quoc Le, and Denny Zhou.
\newblock Chain-of-thought prompting elicits reasoning in large language models.
\newblock \emph{arXiv:2201.11903}, 2022.

\bibitem[Maple(2023)]{maple2023ai}
C.~et~al. Maple.
\newblock The ai revolution: opportunities and challenges for the finance sector.
\newblock \emph{arXiv preprint arXiv:2308.16538}, 2023.

\bibitem[Wu(2023)]{BloombergGPT}
S.~et~al. Wu.
\newblock Bloomberggpt: A genai model for stock prediction and trading.
\newblock \emph{arXiv:2303.17564}, 2023.

\bibitem[et~al.(2019)]{kruse2019artificial}
Kruse et~al.
\newblock Artificial intelligence for the financial services industry: What challenges organizations to succeed, 2019.

\bibitem[Assefa(2020)]{assefa2020generating}
S.~A. et~al. Assefa.
\newblock Generating synthetic data in finance: opportunities, challenges and pitfalls.
\newblock In \emph{Proc. First ACM Int. Conf. AI in Finance}, pages 1--8, 2020.

\bibitem[Ljung(2021)]{ljung2021synthetic}
Mikael Ljung.
\newblock Synthetic data generation for the financial industry using generative adversarial networks, 2021.

\bibitem[Devlin(2019)]{devlin2019bertpretrainingdeepbidirectional}
J.~et~al. Devlin.
\newblock {BERT}: Pre-training of deep bidirectional transformers for language understanding.
\newblock \emph{arXiv:1810.04805}, 2019.

\bibitem[Li(2023{\natexlab{a}})]{li2023large}
Y.~et~al. Li.
\newblock Large language models in finance: A survey.
\newblock In \emph{Proceedings of the fourth {ACM} international conference on {AI} in finance}, pages 374--382, 2023{\natexlab{a}}.

\bibitem[Liu(2021)]{liu2021finbert}
Z.~et~al. Liu.
\newblock Finbert: A pre-trained financial language representation model for financial text mining.
\newblock In \emph{Proceedings of the twenty-ninth international conference on international joint conferences on artificial intelligence}, pages 4513--4519, 2021.

\bibitem[et~al.(2021{\natexlab{a}})]{lora}
Hu~et~al.
\newblock {LORA}: {LOW-RANK ADAPTATION OF LARGE LANGUAGE MODELS}.
\newblock \emph{arXiv:2106.09685}, 2021{\natexlab{a}}.

\bibitem[Xia(2024{\natexlab{b}})]{xia2024understandingperformanceestimatingcost}
Y.~et~al. Xia.
\newblock Understanding the performance and estimating the cost of llm fine-tuning.
\newblock \emph{arXiv:2408.04693 [cs.CL]}, 2024{\natexlab{b}}.

\bibitem[Hancock and Humphrey(1997)]{HANCOCK19971573}
D.~Hancock and D.~B Humphrey.
\newblock Payment transactions, instruments, and systems: A survey.
\newblock \emph{Journal of Banking \& Finance}, 21\penalty0 (11):\penalty0 1573--1624, 1997.

\bibitem[Bryce()]{brycetrends}
C.~et~al. Bryce.
\newblock Trends in large language models: Actors, applications, and impact on cybersecurity.

\bibitem[Srivastava(2024)]{srivastava2024bai}
V.~Srivastava.
\newblock {BAI-Arg LLM} at the finllm challenge task: Earn while you argue-financial argument identification.
\newblock In \emph{Proc. Eighth Financial Technology and Natural Language Processing and the 1st Agent AI for Scenario Planning}, pages 165--173, 2024.

\bibitem[Lakkaraju(2023)]{10.1145/3604237.3626867}
K.~et~al. Lakkaraju.
\newblock {LLMs} for financial advisement: A fairness and efficacy study in personal decision making.
\newblock In \emph{Proc. Fourth ACM Int. Conf. AI in Finance (ICAIF '23)}, pages 100--107, Brooklyn, NY, USA, 2023.
\newblock \doi{10.1145/3604237.3626867}.

\bibitem[Glavina(2024)]{glavina2024ai}
S.~Glavina.
\newblock {AI IN FINANCIAL INDUSTRY: ETHIC ISSUES}, 2024.

\bibitem[Huang et~al.(2024)Huang, Huang, and Catteddu]{huang2024genai}
K.~Huang, J.~Huang, and D.~Catteddu.
\newblock Genai data security.
\newblock In \emph{Generative AI Security: Theories and Practices}, pages 133--162. Springer, 2024.

\bibitem[Zhang(2023)]{zhang2023right}
D.~et~al. Zhang.
\newblock Right to be forgotten in the era of large language models: Implications, challenges, and solutions.
\newblock \emph{arXiv:2307.03941}, 2023.

\bibitem[Rosen(2011)]{rosen2011right}
J.~Rosen.
\newblock The right to be forgotten.
\newblock \emph{Stan. L. Rev. Online}, 64:\penalty0 88, 2011.

\bibitem[Marvin(2023)]{prompt}
Ggaliwango et~al. Marvin.
\newblock Prompt engineering in large language models.
\newblock In \emph{International conference on data intelligence and cognitive informatics}, Singapore, 2023. Springer Nature Singapore.

\bibitem[Wang(2019)]{zeroshot}
Wei et~al. Wang.
\newblock A survey of zero-shot learning: Settings, methods, and applications.
\newblock \emph{{ACM} Transactions on Intelligent Systems and Technology ({TIST})}, 10\penalty0 (2):\penalty0 1--37, 2019.

\bibitem[Brown(2020)]{brown2020language}
Tom~B Brown.
\newblock Language models are few-shot learners.
\newblock \emph{arXiv preprint arXiv:2005.14165}, 2020.

\bibitem[et~al.(2021{\natexlab{b}})]{inductionheads}
Elhage et~al.
\newblock A mathematical framework for transformer circuits.
\newblock Transformer Circuits Thread, 2021{\natexlab{b}}.

\bibitem[et~al.(2022)]{incontextlearning}
Olsson et~al.
\newblock In-context learning and induction heads.
\newblock Transformer Circuits Thread, 2022.

\bibitem[Zhang et~al.(2023)Zhang, Han, Liu, Gao, Zhou, Hu, Yan, Lu, Li, and Qiao]{finetuning}
Renrui Zhang, Jiaming Han, Chris Liu, Peng Gao, Aojun Zhou, Xiangfei Hu, Shilin Yan, Pan Lu, Hongsheng Li, and Yu~Qiao.
\newblock Llama-adapter: Efficient fine-tuning of language models with zero-init attention.
\newblock \emph{arXiv preprint arXiv:2303.16199}, 2023.

\bibitem[Mahabadi et~al.(2021)Mahabadi, Ruder, Dehghani, and Henderson]{tasktuning}
Rabeeh~Karimi Mahabadi, Sebastian Ruder, Mostafa Dehghani, and James Henderson.
\newblock Parameter-efficient multi-task fine-tuning for transformers via shared hypernetworks.
\newblock \emph{arXiv preprint arXiv:2106.04489}, 2021.

\bibitem[Liu et~al.(2022)Liu, Tam, Muqeeth, Mohta, Huang, Bansal, and Raffel]{paramtuning}
Haokun Liu, Derek Tam, Mohammed Muqeeth, Jay Mohta, Tenghao Huang, Mohit Bansal, and Colin~A Raffel.
\newblock Few-shot parameter-efficient fine-tuning is better and cheaper than in-context learning.
\newblock \emph{Advances in Neural Information Processing Systems}, 35:\penalty0 1950--1965, 2022.

\bibitem[Lester et~al.(2021)Lester, Al-Rfou, and Constant]{DBLP:journals/corr/abs-2104-08691}
Brian Lester, Rami Al-Rfou, and Noah Constant.
\newblock The power of scale for parameter-efficient prompt tuning.
\newblock \emph{CoRR}, abs/2104.08691, 2021.

\bibitem[Xu et~al.(2024)Xu, Jain, and Kankanhalli]{xu2024hallucination}
Ziwei Xu, Sanjay Jain, and Mohan Kankanhalli.
\newblock Hallucination is inevitable: An innate limitation of large language models.
\newblock \emph{arXiv preprint arXiv:2401.11817}, 2024.

\bibitem[Maynez et~al.(2020)Maynez, Narayan, Bohnet, and McDonald]{llmfactuality}
Joshua Maynez, Shashi Narayan, Bernd Bohnet, and Ryan McDonald.
\newblock On faithfulness and factuality in abstractive summarization.
\newblock \emph{arXiv preprint arXiv:2005.00661}, 2020.

\bibitem[Patel et~al.(2021)Patel, Bhattamishra, and Goyal]{arellmgoodatmath}
A.~Patel, S.~Bhattamishra, and N.~Goyal.
\newblock Are nlp models really able to solve simple math word problems?
\newblock \emph{CoRR}, abs/2103.07191, 2021.

\bibitem[Nagel et~al.(2021)Nagel, Fournarakis, Amjad, Bondarenko, van Baalen, and Blankevoort]{DBLP:journals/corr/abs-2106-08295}
Markus Nagel, Marios Fournarakis, Rana~Ali Amjad, Yelysei Bondarenko, Mart van Baalen, and Tijmen Blankevoort.
\newblock A white paper on neural network quantization.
\newblock \emph{CoRR}, abs/2106.08295, 2021.

\bibitem[Gholami et~al.(2021)Gholami, Kim, Dong, Yao, Mahoney, and Keutzer]{DBLP:journals/corr/abs-2103-13630}
Amir Gholami, Sehoon Kim, Zhen Dong, Zhewei Yao, Michael~W. Mahoney, and Kurt Keutzer.
\newblock A survey of quantization methods for efficient neural network inference.
\newblock \emph{CoRR}, abs/2103.13630, 2021.

\bibitem[Bengio(2013)]{DBLP:journals/corr/abs-1305-2982}
Yoshua Bengio.
\newblock Estimating or propagating gradients through stochastic neurons.
\newblock \emph{CoRR}, abs/1305.2982, 2013.

\bibitem[Chen(2021)]{finqa}
Zhiyu et~al. Chen.
\newblock Finqa: A dataset of numerical reasoning over financial data.
\newblock \emph{arXiv preprint arXiv:2109.00122}, 2021.

\bibitem[Chen(2022)]{convfinqa}
Zhiyu et~al. Chen.
\newblock Convfinqa: Exploring the chain of numerical reasoning in conversational finance question answering.
\newblock \emph{arXiv preprint arXiv:2210.03849}, 2022.

\bibitem[Li(2023{\natexlab{b}})]{gpt-on-fin-datasets}
Xianzhi et~al. Li.
\newblock Are chatgpt and gpt-4 general-purpose solvers for financial text analytics? a study on several typical tasks.
\newblock \emph{arXiv preprint arXiv:2305.05862}, 2023{\natexlab{b}}.

\bibitem[{Fintool team}()]{fintool}
{Fintool team}.
\newblock Fintool.
\newblock \url{https://fintool.com/}.

\bibitem[Graciano(2024)]{graciano2024credit}
R.~Graciano.
\newblock Intuit credit karma scales genai-first experiences.
\newblock \emph{Credit Karma}, July 2024.

\bibitem[Mai(2024)]{stockgpt}
D.~Mai.
\newblock Stockgpt: A genai model for stock prediction and trading.
\newblock \emph{arXiv preprint arXiv:2404.05101}, 2024.

\bibitem[Lopez-Lira and Tang(2023)]{Lopez_Lira_2023}
A.~Lopez-Lira and Y.~Tang.
\newblock Can chatgpt forecast stock price movements? return predictability and large language models.
\newblock \emph{SSRN Electronic Journal}, 2023.

\bibitem[Fatouros(2024)]{can-llm-beat-wallstreet}
Georgios et~al. Fatouros.
\newblock Can large language models beat wall street? unveiling the potential of ai in stock selection.
\newblock \emph{arXiv preprint arXiv:2401.03737}, 2024.

\bibitem[Li(2023{\natexlab{c}})]{li2023multimodal}
L.~et~al. Li.
\newblock Multimodal gen-ai for fundamental investment research.
\newblock \emph{arXiv preprint arXiv:2401.06164}, 2023{\natexlab{c}}.

\bibitem[Pu et~al.(2023)Pu, Gao, and Wan]{summarization-is-dead}
Xiao Pu, Mingqi Gao, and Xiaojun Wan.
\newblock Summarization is (almost) dead.
\newblock \emph{arXiv preprint arXiv:2309.09558}, 2023.

\bibitem[Colombo(2024)]{colombo2024saullm7b}
P.~et~al. Colombo.
\newblock {SaulLM-7B}: A pioneering large language model for law.
\newblock \emph{arXiv preprint arXiv:2403.03883}, 2024.

\bibitem[jpm(2024)]{jpmorganResearch}
Jpmorgan chase leads ai revolution in finance with launch of llm suite.
\newblock \emph{Forbes}, Jul 2024.

\bibitem[{Salesforce}()]{salesforceEinsteinAI}
{Salesforce}.
\newblock Einstein ai assistant.

\bibitem[{Hubspot}()]{hubspotChatbotBuilder}
{Hubspot}.
\newblock Hubspot chatbot builder.

\bibitem[{Zendesk}()]{zendeskChatbotBuilder}
{Zendesk}.
\newblock Zendesk chatbot builder.

\bibitem[Greene-Lewis(2023)]{turbotax}
Lisa Greene-Lewis.
\newblock Transformative tax preparation: Gen-ai powered intuit assist, September 2023.

\bibitem[Brown(2024)]{brown2024decision}
R.~Brown.
\newblock Mastercard launches gpt-like ai model to help banks detect fraud.
\newblock \emph{CNBC}, Feb 2024.

\bibitem[Cao et~al.(2024)Cao, Chen, Pei, Dimino, Ausiello, Kumar, Subbalakshmi, and Ndiaye]{risklabs}
Yupeng Cao, Zhi Chen, Qingyun Pei, Fabrizio Dimino, Lorenzo Ausiello, Prashant Kumar, K.~P. Subbalakshmi, and Papa~Momar Ndiaye.
\newblock Risklabs: Predicting financial risk using large language model based on multi-sources data.
\newblock \emph{arXiv preprint arXiv:2404.07452}, 2024.

\end{thebibliography}
\vspace{12pt}
\color{red}

\end{document}